\documentclass{article}
\usepackage[preprint,nonatbib]{neurips_2019}

\usepackage[utf8]{inputenc} 
\usepackage[T1]{fontenc}    
\usepackage{hyperref}       
\usepackage{url}            
\usepackage{booktabs}       
\usepackage{amsfonts}       
\usepackage{nicefrac}       
\usepackage{microtype}      

\usepackage{times}
\usepackage{wrapfig}
\usepackage{latexsym}
\usepackage{graphicx}
\usepackage{algorithm2e}
\usepackage{amsmath}
\usepackage{placeins}
\usepackage{caption}
\usepackage{subcaption}
\usepackage{footnote}
\usepackage{amssymb}
\usepackage{lipsum}
\usepackage{appendix}
\usepackage{colortbl}
\usepackage{multirow}
\usepackage{algorithmicx}
\usepackage{bm}
\makesavenoteenv{tabular}
\usepackage[numbers]{natbib}

\usepackage[noend]{algpseudocode}

\definecolor{gray}{RGB}{87, 87, 87}
\definecolor{red}{RGB}{173, 35, 35}
\definecolor{blue}{RGB}{42, 75, 215}
\definecolor{green}{RGB}{29, 105, 20}
\definecolor{brown}{RGB}{129, 74, 25}
\definecolor{purple}{RGB}{129, 38, 192}
\definecolor{cyan}{RGB}{41, 208, 208}
\definecolor{yellow}{RGB}{189, 167, 0}
\definecolor{Red}{rgb}{0.68, 0.05, 0.0}
\definecolor{Blue}{rgb}{0.0, 0.0, 0.61}
\definecolor{Blue1}{RGB}{214, 235, 245}
\definecolor{Blue2}{RGB}{235, 245, 250}
\definecolor{lime}{RGB}{60,179,113}
\definecolor{peach}{RGB}{255, 242, 230}

\newcommand{\specialcell}[2][c]{\begin{tabular}[#1]{@{}c@{}}#2\end{tabular}}
\newcommand{\norm}[1]{\left\lVert#1\right\rVert}

\title{Structure Learning for Neural Module Networks}

\author{Vardaan Pahuja$^{\dag\S}$\thanks{Corresponding author: Vardaan Pahuja <vardaanpahuja@gmail.com>} \qquad Jie Fu$^{\dag\ddag\diamond}$ \qquad Sarath Chandar$^{\dag\S}$ \qquad Christopher J. Pal$^{\dag\ddag\mathparagraph\star}$\\~\\
$^{\dag}$Mila \qquad $^{\S}$Université de Montréal \qquad $^\ddag$Polytechnique Montréal \qquad $^{\diamond}$IVADO\\
$^{\mathparagraph}$Element AI \qquad $^{\star}$Canada CIFAR AI Chair}

\date{}

\begin{document}
\maketitle
\begin{abstract}

Neural Module Networks, originally proposed for the task of visual question answering, are a class of neural network architectures that involve human-specified neural modules, each designed for a specific form of reasoning. In current formulations of such networks only the parameters of the neural modules and/or the order of their execution is learned. In this work, we further expand this approach and also learn the underlying internal structure of modules in terms of the ordering and combination of simple and elementary arithmetic operators. Our results show that one is indeed able to simultaneously learn both internal module structure and module sequencing without extra supervisory signals for module execution sequencing. With this approach, we report performance comparable to models using hand-designed modules.

\end{abstract}

\section{Introduction}

Designing general purpose reasoning modules is one of the central challenges in artificial intelligence. Neural Module Networks \cite{andreas2016neural} were introduced as a general purpose visual reasoning architecture and have been shown to work well for the task of visual question answering \cite{VQA, malinowski2014multi, ren2015image, ren2015exploring}.
They use dynamically composable modules which are then assembled into a layout based on syntactic parse of the question. The modules take as input the images or the attention maps\footnote{An attention map denotes a $H\times W\times 1$ tensor which assigns a saliency score to the convolutional features in the spatial dimension.} and return attention maps or labels as output. In \cite{hu2017learning}, the layout prediction is relaxed by learning a layout policy with a sequence-to-sequence RNN. This layout policy is jointly trained along with the parameters of the modules. The model proposed in \cite{hu2018explainable} avoids the use of reinforcement learning to train the layout predictor, and uses soft program execution to learn both layout and module parameters jointly.


A fundamental limitation of these previous modular approaches to visual reasoning is that the modules need to be hand-specified. This might not be feasible when one has limited knowledge of the kinds of questions or associated visual reasoning required to solve the task. In this work, we present an approach to learn the module structure, along with the parameters of the modules in an end-to-end differentiable training setting. Our proposed model, Learnable Neural Module Network (LNMN), learns the structure of the module, the parameters of the module, and the way to compose the modules based on just the regular task loss. Our results show that we can learn the structure of the modules automatically and still perform comparably to hand-specified modules. We want to highlight the fact that our goal in this paper is not to beat the performance of the hand-specified modules since they are specifically engineered for the task. Instead, our goal is to explore the possibility of designing general-purpose reasoning modules in an entirely data-driven fashion.




\pagebreak

\section{Background}\label{backgrnd}

In this section, we describe the working of the \textit{Stack-NMN} model \cite{hu2018explainable} as our proposed LNMN model uses this as the base model.
The \textit{Stack-NMN} model is an end-to-end differentiable model for the task of Visual Question Answering and Referential Expression Grounding \cite{rohrbach2016grounding}. It addresses a major drawback of prior visual reasoning models in the literature that compositional reasoning is implemented without the need of supervisory signals for composing the layout at training time. It consists of several hand-specified modules (namely Find, Transform, And, Or, Filter, Scene, Answer, Compare and NoOp) which are parameterized, differentiable, and implement common routines needed in visual reasoning and learns to compose them without strong supervision. The implementation details of these modules are given in Appendix~\ref{sec:hand_modules} (see Table~\ref{tab:modules}). The different sub-components of the Stack-NMN model are described below.

\subsection{Module Layout Controller}
The structure of the controller is similar to the one proposed in \cite{hudson2018compositional}. The controller first encodes the question using a bi-directional LSTM \cite{hochreiter1997long}. Let $[\bm{h_1}, \bm{h_2}, ..., \bm{h_S}]$ denote the output of Bi-LSTM at each time-step of the input sequence of question words. Let $\bm{q}$ denote the concatenation of final hidden state of Bi-LSTM during the forward and backward passes. $\bm{q}$ can be considered as the encoding of the entire question. The controller executes the modules iteratively for $T$ times. At each time-step, the updated query representation $\bm{u}$ is obtained as:
\begin{wrapfigure}[28]{r}{0.5\textwidth}
\begin{minipage}{0.5\textwidth}
\begin{algorithm}[H]
 \KwData{Question (string), Image features ($\mathcal{I}$)}
 Encode the input question into $d$-dimensional sequence $[\bm{h_1}, \bm{h_2}, ..., \bm{h_S}]$ using Bidirectional LSTM.\\
 $A^{(0)} \gets$ Initialize the memory stack $(A; p)$ with uniform image attention and set the stack pointer $p$ to point at the bottom of the stack (one-hot vector with $1$ in the $1^{st}$ dim.).
 
 \For{each time-step t = 0, 1, ...., (T-1)}{
  $\bm{u} = \bm{W_2}[\bm{W_1^{(t)}} \bm{q} + \bm{b_1}; \bm{c_{t-1}}] + \bm{b_2}$\;
  $\bm{w^{(t)}} = softmax(MLP(\bm{u}; \theta_{MLP}))$\;
  $cv_{t,s} = softmax(W_3(\bm{u} \odot \bm{h_s}))$\;
  $\bm{c_t} = \sum_{s=1}^S cv_{t,s} \cdot \bm{h_s}$\\
  \For{every module $m \in M$}{
  Produce updated stack and stack pointer: $(A_m^{(t)}, p_m^{(t)}) = \textrm{run-module}(m, A^{(t)}, p^{(t)}, \bm{c_t}, \mathcal{I})$\;
  }
  $A^{(t+1)} = \sum_{m \in M} A_m^{(t)} \cdot w_m^{(t)}$\;
  $p^{(t+1)} = softmax(\sum_{m \in M} p_m^{(t)} \cdot w_m^{(t)})$
 }
 \caption{Operation of Module Layout Controller and Memory Stack.}
 \label{algo:stacknmn}
\end{algorithm}
\end{minipage}
\end{wrapfigure}
$$\bm{u} = \bm{W_2}[\bm{W_1^{(t)}} \bm{q} + \bm{b_1};\bm{ c_{t-1}}] + \bm{b_2}$$
where $\bm{W_1^{(t)}} \in \mathbb{R}^{d \times d}$, $\bm{W_2} \in \mathbb{R}^{d \times 2d}$, $\bm{b_1} \in \mathbb{R}^{d}$, $\bm{b_2} \in \mathbb{R}^{d}$ are controller parameters. $\bm{c_{t-1}}$ is the textual parameter from the previous time step. The controller has two outputs viz. the textual parameter at step $t$ (denoted by $\bm{c_t}$) and the attention on each module (denoted by vector $\bm{w^{(t)}}$). The controller first predicts an attention $cv_{t,s}$ on each of the words of the question and then uses this attention to do a weighted average over the outputs of the Bi-LSTM.
$$cv_{t,s} = softmax(\bm{W_3}(\bm{u} \odot \bm{h_s}))$$
$$\bm{c_t} = \sum_{s=1}^S cv_{t,s} \cdot \bm{h_s}$$
where, $\bm{W_3} \in \mathbb{R}^{1 \times d}$ is another controller parameter. The attention on each module $\bm{w^{(t)}}$ is obtained by feeding the query representation at each time-step to a Multi-layer Perceptron (MLP).
$$\bm{w^{(t)}} = softmax(MLP(\bm{u}; \theta_{MLP}))$$

\subsection{Operation of Memory Stack for storing attention maps}
In order to answer a visual reasoning question, the model needs to execute modules in a tree-structured layout. In order to facilitate this sort of compositional behavior, a differentiable memory pool to store and retrieve intermediate attention maps is used. A memory stack (with length denoted by $L$) stores $H \times W$ dimensional attention maps, where $H$ and $W$ are the height and width of image feature maps respectively. Depending on the number of attention maps required as input by the module, it pops them from the stack and later pushes the result back to the stack. The model performs soft module execution by executing all modules at each time step. The updated stack and stack pointer at each subsequent time-step are obtained by a weighted average of those corresponding to each module using the weights $\bm{w^{(t)}}$ predicted by the module controller. This is illustrated by the equations below:
\begin{align*}
(A_m^{(t)}, p_m^{(t)}) &= \textrm{run-module}(m, A^{(t)}, p^{(t)})\\
A^{(t+1)} &= \sum_{m \in M} A_m^{(t)} \cdot w_m^{(t)}\\
p^{(t+1)} &= softmax(\sum_{m \in M} p_m^{(t)} \cdot w_m^{(t)})
\end{align*}
Here, $A_m^{(t)}$ and $p_m^{(t)}$ denote the stack and stack pointer respectively, after executing module $m$ at time-step $t$. $A^{(t)}$ and $p^{(t)}$ denote the stack and stack pointer obtained after the weighted average of those corresponding to all modules at previous time-step $(t-1)$. The working of module layout controller and its interfacing with memory stack is illustrated in Algorithm~\ref{algo:stacknmn}. The implementation details of operation of the stack are shown in Appendix (see Algorithm~\ref{algo:module}).

\subsection{Final Classifier}
At each time-step of module execution, the weighted average of output of the \textit{Answer} modules is called memory features (denoted by $f_{mem}^{(t)}=\sum_{m\in\textrm{ans. module}} o_m^{(t)} w_m^{(t)}$). Here, $o_m^{(t)}$ denotes the output of module $m$ at time $t$. The memory features are given as one of the inputs to the \textit{Answer} modules at the next time-step. The memory features at the final time-step are concatenated with the question representation, and then fed to an MLP to obtain the logits.
\section{Learnable Neural Module Networks}\label{proposed_model}

In this section, we introduce Learnable Neural Module Networks (LNMNs) for visual reasoning, which extends Stack-NMN. However, the modules in LNMN are not hand-specified. Rather, they are generic modules as specified below.

\subsection{Structure of the Generic Module} \label{sec:generic_module}
\begin{figure*}[htbp]
\centering\includegraphics[scale=0.3]{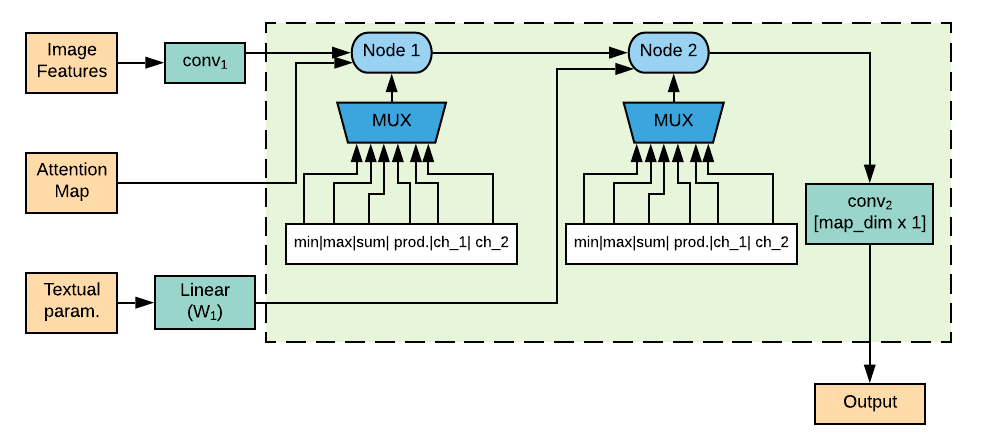}
\caption{Attention Module schematic diagram (3 inputs).\label{fig:cell_structure_3_input}}
\end{figure*}

The \textit{cell} (see Figure~\ref{fig:cell_structure_3_input}) denotes a generic module, which we suppose can span all the required modules for a visual reasoning task. Each cell contains a certain number of nodes. The function of a node (denoted by $O$) is to perform a weighted sum of outputs of different arithmetic operations applied on the input feature maps $\bm{x_1}$ and $\bm{x_2}$. Let ${\bm{\alpha}}^{'} = \sigma(\bm{\alpha})$ denote the output of softmax function applied to the vector $\bm{\alpha}$ such that
\begin{align*}
O(\bm{x_1},\bm{x_2}) &= \alpha^{'}_1 * min(\bm{x_1}, \bm{x_2})
+ \alpha^{'}_2 * max(\bm{x_1}, \bm{x_2}) + \alpha^{'}_3 * (\bm{x_1} + \bm{x_2})\\
&+\alpha^{'}_4 * (\bm{x_1} \odot \bm{x_2}) + \alpha^{'}_5 * choose_1 (\bm{x_1}, \bm{x_2})
+ \alpha^{'}_6 * choose_2 (\bm{x_1}, \bm{x_2})
\end{align*}
All of the above operations (\textit{min}, \textit{max}, $+$, $\odot$) are element-wise operations. The last two non-standard functions are defined as: $choose_1(\bm{x_1}, \bm{x_2}) = \bm{x_1} \textrm{ and } choose_2(\bm{x_1}, \bm{x_2}) = \bm{x_2}$.

We consider two broad kinds of modules: (i) \textit{Attention modules} which output an attention map (ii) \textit{Answer modules} which output memory features to be stored in the memory. Among each of these two categories, there is a finer categorization:
\subsubsection{Generic Module with 3 inputs}
This module type receives 3 inputs (i.e. image features, textual parameter, and a single attention map) and produces a single output. The first node receives input from the image feature ($\mathcal{I}$) and the attention map (popped from the memory stack). The second node receives input from the textual parameter followed by a linear layer ($W_1c_{txt}$), and the output of the first node.

\subsubsection{Generic Module with 4 inputs}
This module type receives 4 inputs (i.e. image features, textual parameter and two attention maps) and produces a single output. The first node receives the two attention maps, each of which are popped from the memory stack, as input. The second node receives input from the image features along with the output of the first node. The third node receives input from the textual parameter followed by a linear layer, and the output of the second node.

For the \textit{Attention} modules, the output of the final node is converted into a single-channel attention map using a $1\times 1$ convolutional layer. For the \textit{Answer} modules, the output of the final node is summed over spatial dimensions, and the resulting feature vector is concatenated with memory features of previous time-step and textual parameter features, fed to a linear layer to output memory features. The schematic diagrams of the \textit{Attention} module with four inputs and \textit{Answer} modules are given in the Appendix~\ref{sec:module_diag_appendix} (see Figures~\ref{fig:cell_structure_4_input}, ~\ref{fig:cell_structure_3_input_answer}, ~\ref{fig:cell_structure_4_input_answer}).

\subsection{Overall structure}
The structure of our end-to-end model extends \textit{Stack-NMN} in that we specify each module in terms of the generic module (defined in Section \ref{sec:generic_module}). We experiment with three model ablations in terms of number of modules for each type being used. See Table~\ref{tab:n_modules_type} for details\footnote{1 NoOp module is included by default in all ablations.}.
We train the module structure parameters (denoted by $\bm{\alpha} = {\left\{\alpha_i^{m,k}\right\}}_{i=1}^{6}$ for $k^{th}$ node of module $m$) and the weight parameters ($\mathcal{W}$) by adopting alternating gradient descent steps in architecture and weight spaces respectively.
For a particular epoch, the gradient step in weight space is performed on each training batch, and the gradient step in architecture space is performed on a batch randomly sampled from the validation set.
\begin{wrapfigure}[18]{o}{0.5\textwidth}
\begin{algorithm}[H]
\DontPrintSemicolon
\While{not converged}{
	1. Update weights $\mathcal{W}$ by descending $\nabla_{\bm{w}} \Big[\mathcal{L}_{train}(\mathcal{W}, \bm{\alpha}) - \frac{\lambda_w}{T} \sum\limits_{t=1}^T \mathcal{H}(\bm{w^{(t)}})\Big]$\;
	2. Update architecture $\bm{\alpha}$ by descending\\
	$\nabla_{\bm{\alpha}} \Big[\mathcal{L}_{val}(\mathcal{W}, \bm{\alpha})-\lambda_{op} \sum\limits_{m=1}^M\sum\limits_{k=1}^p \frac{\norm{\bm{\sigma}(\bm{\alpha}^{m,k})}_2}{\norm{\bm{\sigma}(\bm{\alpha}^{m,k})}_1}\Big]$\;
}
\caption{Training Algorithm for LNMN Modules. Here, $\bm{\alpha}$ denotes the collection of module network parameters i.e. ${\left\{\alpha_i^{m,k}\right\}}_{i=1}^{6}$ for $k^{th}$ node of module $m$, $\mathcal{W}$ denotes the collection of weight parameters of modules and all other non-module parameters.}
\label{algo:pseudocode}
\end{algorithm}
\end{wrapfigure}
This is done to ensure that we find an architecture corresponding to the modules which has a low validation loss on the updated weight parameters. This is inspired by the technique used in \cite{liu2018darts} to learn monolithic architectures like CNNs and RNNs in terms of basic building blocks (or \textit{cells}).
Algorithm~\ref{algo:pseudocode} illustrates the training algorithm. Here, $\mathcal{L}_{train}(\mathcal{W}, \bm{\alpha})$ and $\mathcal{L}_{val}(\mathcal{W}, \bm{\alpha})$ denote the training loss and validation loss on the combination of parameters $(\mathcal{W}, \bm{\alpha})$ respectively. For the gradient step on the training batch, we add an additional loss term to initially maximize the entropy of $\bm{w^{(t)}}$ and gradually anneal the regularization coefficient ($\lambda_w$) to the opposite sign (which minimizes the entropy of $\bm{w^{(t)}}$ towards the end of training). The value of $\lambda_w$ varies linearly from $1.0$ to $0.0$ in the first 20 epochs and then steeply decreases to $-1.0$ in next 10 epochs. The trend of variation of $\lambda_w$ is shown in Appendix (see Figure~\ref{fig:reg_coeff_plot}). 
For the gradient steps in the architecture space, we add an additional loss term ($\frac{l^2}{l^1} = \frac{\norm{\bm{\sigma}(\bm{\alpha})}_2}{\norm{\bm{\sigma}(\bm{\alpha})}_1}$) \cite{hurley2009comparing} to encourage the sparsity of module structure parameters ($\bm{\alpha}$) after the softmax activation. 



\section{Experiments}\label{expt}
We train our model on the CLEVR visual reasoning task. CLEVR \cite{johnson2017clevr} is a synthetic dataset for visual reasoning containing around 700K examples, and has become the standard benchmark to test visual reasoning models. It contains questions that test visual reasoning abilities such as counting, comparing, logical reasoning based on 3D shapes like cubes, spheres, and cylinders of varied shades.
A typical example question and image pair from this dataset is given in Appendix (see Figure~\ref{fig:clevr_img}).
The results on CLEVR test set are reported in Table~\ref{tab:fullclevr}. Some ablations of the model are shown in Table~\ref{tab:ablations}. We use the pre-trained CLEVR model to fine-tune the model on CLEVR-Humans dataset. The latter is a dataset of challenging human-posed questions based on a much larger vocabulary on the same CLEVR images. The corresponding results are shown in Table~\ref{tab:fullclevr} (see last column). In addition, we experiment on VQA v1 \cite{VQA} and VQA v2 \cite{balanced_vqa_v2} which are VQA datasets containing natural images. The results for VQA v1 and VQA v2 are shown in Table~\ref{tab:results_vqa}.


\begin{table*}[!ht]
\centering
\small
\resizebox{0.98\linewidth}{!}{
\begin{tabular}{l@{}cccccc|cc}
\toprule
Model & \textbf{CLEVR} & Count & Exist & Compare & Query & Compare & \textbf{CLEVR} \\ 
 & \textbf{Overall}            &           &          & Numbers & Attribute & Attribute & \textbf{Humans} \\
\midrule
Human \cite{johnson2017inferring} & 92.6 & 86.7 & 96.6 & 86.5 & 95.0 & 96.0 & - \\
Q-type baseline \cite{johnson2017inferring} & 41.8 & 34.6 & 50.2 & 51.0 & 36.0 & 51.3 & - \\
LSTM \cite{johnson2017inferring} & 46.8 & 41.7 & 61.1 & 69.8 & 36.8 & 51.8 & 36.5 \\
CNN+LSTM \cite{johnson2017inferring} & 52.3 & 43.7 & 65.2 & 67.1 & 49.3 & 53.0 & 43.2 \\
CNN+LSTM+SA+MLP \cite{johnson2017clevr} & 73.2 & 59.7 & 77.9 & 75.1 & 80.9 & 70.8 & 57.6 \\
\hline
N2NMN* \cite{hu2017learning} & 83.7 & 68.5 & 85.7 & 84.9 & 90.0 & 88.7 & - \\
PG+EE (700K prog.)* \cite{johnson2017inferring} & 96.9 & 92.7 & 97.1 & 98.7 & 98.1 & 98.9 & - \\
CNN+LSTM+RN$^{\ddag}$ \cite{santoro2017simple} & 95.5 & 90.1 & 97.8 & 93.6 & 97.9 & 97.1 & - \\
CNN+GRU+FiLM \cite{DBLP:journals/corr/abs-1709-07871} & 97.7 & 94.3 & 99.1 & 96.8 & 99.1 & 99.1 & 75.9\\
MAC \cite{hudson2018compositional} & 98.9 & 97.1 & 99.5 & 99.1 & 99.5 & 99.5 & 81.5\\
TbD \cite{Mascharka2018TransparencyBD} & 99.1 & 97.6 & 99.2 & 99.4 & 99.5 & 99.6 & -\\
\midrule
\vspace{1ex}
Stack-NMN (9 mod.)$^{\dag}$\cite{hu2018explainable} & 91.41 & 81.78 & 95.78 & 85.23 & 95.45 & 95.68 & 68.06\\
\rowcolor{Blue1} LNMN (9 modules) & 89.88 & 84.28 & 93.74 & 89.63 & 89.64 & 94.84 & 66.35\\
\rowcolor{Blue1} LNMN (11 modules) & 90.52 & 84.91 & 95.21 & 91.06 & 90.03 & 94.97 & 65.68\\
\rowcolor{Blue1} LNMN (14 modules) & 90.42 & 84.79 & 95.52 & 90.52 & 89.73 & 95.26 & 65.86\\
\bottomrule
\end{tabular}}
\caption{CLEVR and CLEVR-Humans Accuracy by baseline methods and our models. (*) denotes use of extra supervision through program labels. ($^{\ddag}$) denotes training from raw pixels. $^{\dag}$ Accuracy figures for our implementation of Stack-NMN.}
\label{tab:fullclevr}
\end{table*}
\begin{table*}[!ht]
\footnotesize
\centering
\resizebox{0.95\linewidth}{!}{
\begin{tabular}{c|c|cccccc}
\toprule
Model   & Overall & Count & Exist & \specialcell{Compare\\number} & \specialcell{Query\\attribute} & \specialcell{Compare\\Attribute} \\ \midrule 
\rowcolor{peach} \specialcell{Original setting\\$(T=5, L=5, \textrm{map}\_\textrm{dim}=384)$} & 89.78 & 84.54 & 93.46 & 88.70 & 89.59 & 94.87 \\ \midrule
\specialcell{Use hard-max for operation weights\\(for inference only)\\$(T=5, L=5, \textrm{map}\_\textrm{dim}=384)$} & 87.99 & 81.53 & 94.11 & 87.70 & 88.27 & 91.55  \\ \hline
$T=9, L=9, \textrm{map}\_\textrm{dim}=256$ & 89.96 & 84.03 & 93.45 & 89.98 & 90.75 & 93.10 \\ \hline 
\specialcell{Concatenate all inputs\\followed by conv. layer}  & 47.03 & 42.5 & 61.15 & 68.64 & 38.06 & 49.43 \\ \bottomrule
\end{tabular}}
\caption{Model Ablations for LNMN (CLEVR Validation set performance). The term `$\textrm{map}\_\textrm{dim}$' refers to the dimension of feature representation obtained at the input or output of each node of cell.}
\label{tab:ablations}
\end{table*}

\noindent\begin{minipage}{.55\textwidth}
\small
\resizebox{0.95\linewidth}{!}{
\begin{tabular}{c|cccc}
\toprule
Model                  & \specialcell{Attn.\\mod.\\(3 inp.)} & \specialcell{Attn.\\mod.\\(4 inp.)} & \specialcell{Ans.\\mod.\\(3 inp.)} & \specialcell{Ans.\\mod.\\(4 inp.)} \\ \midrule
\specialcell{LNMN (9)}  & 4                           & 2                           & 1                        & 1                        \\
\specialcell{LNMN (11)} & 4                           & 2                           & 2                        & 2                        \\
\specialcell{LNMN (14)} & 5                           & 4                           & 2                        & 2 \\        
\bottomrule
\end{tabular}}
\captionof{table}{Number of modules of each type for different model ablations.}
\label{tab:n_modules_type}
\end{minipage}
\noindent\begin{minipage}{.45\textwidth}
\small
\resizebox{0.95\linewidth}{!}{
\begin{tabular}{ccc}
\toprule
Model            & VQA v2 & VQA v1\\ \midrule
Stack-NMN        & 58.23 & 59.84 \\
LNMN (9 modules) & 54.85 & 57.67  \\ \bottomrule
\end{tabular}}
\captionof{table}{Test Accuracy on Natural Image VQA datasets}
\label{tab:results_vqa}
\end{minipage}

The detailed accuracy for each question sub-type for the VQA datasets is given in Appendix~\ref{sec:acc_vqa} (see Tables~\ref{tab:results_vqa_1} and \ref{tab:results_vqa_2}). We use Adam \cite{Kingma2014} as the optimizer for the weight parameters with a learning rate of $1\mathrm{e}{-4}$, $(\beta_1, \beta_2) = (0.9, 0.999)$ and no weight decay.
For the module network parameters, we use the same optimizer with a different learning rate $3\mathrm{e}{-4}$, $(\beta_1, \beta_2) = (0.5, 0.999)$ and a weight decay of $1\mathrm{e}{-3}$. The value of $\lambda_{op}$ is set to $1.0$.

\subsection{Results}\label{results}
The comparison of CLEVR overall accuracy shows that our model (LNMN (9 modules)) receives only a slight dip ($1.53\%$) compared to the Stack-NMN model. We also experiment with other variants of our model in which we increase the number of \textit{Answer} modules (LNMN (11 modules)) and/or \textit{Attention} modules (LNMN (14 modules)). The LNMN (11 modules) model performs better than the other two ablations ($0.89\%$ accuracy drop w.r.t. the Stack-NMN model). For the `Count' and `Compare Numbers' sub-category of questions, all of the 3 variants perform consistently better than the Stack-NMN model. In case of CLEVR-Humans dataset, the accuracy drop is a modest $1.71\%$. Even for the natural image VQA datasets, our approach has comparable results with the Stack-NMN model.
The results clearly show that the modules learned by our model (in terms of elementary arithmetic operations) perform approximately as well as the ones specified in the Stack-NMN model (that contains hand-designed modules which were tailor-made for the CLEVR dataset). The results from the ablations in Table~\ref{tab:ablations} show that a naive concatenation of all inputs to a module (or \textit{cell}) results in a poor performance (around 47 $\%$). Thus, the structure we propose to fuse the inputs plays a key role in model performance. When we replace the $\bm{\alpha}$ vector for each node by a one-hot vector during inference, the drop in accuracy is only $1.79\%$ which shows that the learned distribution over operation weights peaks over a specific operation which is desirable.


\begin{table*}[!ht]
\small
\centering
\resizebox{0.95\linewidth}{!}{
\begin{tabular}{c|c|cccccc}
\toprule
\textbf{Module ID} & Module type & \textbf{min} & \textbf{max} & \textbf{sum} & \textbf{product} & \textbf{choose\_1} & \textbf{choose\_2} \\ \midrule
0                  & Attn. (3 input) & 6.3  $\mathrm{e} {4}$ & 2.7  $\mathrm{e} {4}$	& 3.6  $\mathrm{e} {4}$ & 1.1  $\mathrm{e} {5}$	& 5.1  $\mathrm{e} {4}$ & 1.6  $\mathrm{e} {4}$             \\
1                  & Attn. (3 input) & 4.4  $\mathrm{e} {4}$ & 1.8  $\mathrm{e} {4}$	& 6.2  $\mathrm{e} {4}$ & 1.4  $\mathrm{e} {4}$ & 2.8  $\mathrm{e} {4}$ & 1.7  $\mathrm{e} {5}$             \\
2                  & Attn. (3 input) & 7.0  $\mathrm{e} {4}$ & 3.3  $\mathrm{e} {4}$	& 3.8  $\mathrm{e} {4}$ & 1.1  $\mathrm{e} {5}$ & 5.2  $\mathrm{e} {4}$ & 1.5  $\mathrm{e} {4}$              \\
3                  & Attn. (3 input) & 8.6  $\mathrm{e} {3}$ & 6.2  $\mathrm{e} {4}$	& 1.7  $\mathrm{e} {4}$ & 1.8  $\mathrm{e} {4}$	& 4.7  $\mathrm{e} {4}$ & 3.0  $\mathrm{e} {4}$              \\ \hline
4                  & Attn. (4 input) & 4.5  $\mathrm{e} {4}$ & 3.2  $\mathrm{e} {4}$	& 7.6  $\mathrm{e} {4}$ & 1.7  $\mathrm{e} {4}$ & 3.6  $\mathrm{e} {4}$ & 2.1  $\mathrm{e} {5}$             \\
5                  & Attn. (4 input) & 1.1  $\mathrm{e} {5}$ & 5.6  $\mathrm{e} {5}$	& 2.3  $\mathrm{e} {5}$ & 8.5  $\mathrm{e} {3}$	& 2.8  $\mathrm{e} {4}$ & 1.8  $\mathrm{e} {5}$             \\ \hline
6                  & Ans. (3 input) & 2.1  $\mathrm{e} {6}$ & 4.3  $\mathrm{e} {6}$	& 4.4  $\mathrm{e} {6}$	& 8.3  $\mathrm{e} {6}$ & 2.3  $\mathrm{e} {6}$ & 4.9  $\mathrm{e} {5}$            \\ \hline
7                  & Ans. (4 input) & 1.2  $\mathrm{e} {5}$ & 5.8  $\mathrm{e} {4}$	& 1.7  $\mathrm{e} {5}$ & 5.2  $\mathrm{e} {3}$ & 1.0  $\mathrm{e} {5}$	& 4.5  $\mathrm{e} {5}$          \\ \bottomrule
\end{tabular}}
\caption{Analysis of gradient attributions of $\alpha$ parameters corresponding to each module (LNMN (9 modules)), summed across all examples of CLEVR validation set.}
\label{tab:clevr_ig}
\end{table*}
\subsection{Measuring the sensitivity of modules}
We use an attribution technique called Integrated Gradients \cite{sundararajan2017axiomatic} to study the impact of module structure parameters (denoted by ${\left\{\alpha_i^{m,k}\right\}}_{i=1}^{6}$ for $k^{th}$ node of module $m$) on the probability distribution in the last layer of LNMN model. Let $\mathcal{I}_j$ and $\bm{q}_j$ denote the (image, question) pairs for the $j^{th}$ example. Let $F(\mathcal{I}_j, \bm{q}_j, \bm{\alpha})$ denote the function that assigns the probability corresponding to the correct answer index in the softmax distribution.  Here, $\alpha^{m,k}_i$ denotes the module network parameter for the $i^{th}$ operator in $k^{th}$ node of module $m$. Then, the attribution of $[\alpha^m_1, \alpha^m_2, \alpha^m_3, \alpha^m_4, \alpha^m_5, \alpha^m_6]$ (summed across all nodes $k = 1,...,p$ for a particular module $m$ and over all examples) is defined as:
\begin{align*}
&\textrm{IG}(\alpha^m_i) = \sum_{j=1}^{N}\sum_{k=1}^{p}\bigg[(\alpha^{m,k}_i - (\alpha^{m,k}_i)^{'})\times \int_{\xi=0}^{1} \frac{\partial F(\mathcal{I}_j, \bm{q}_j, (1-\xi)\times (\alpha^{m,k}_i)^{'}+ \xi \times \alpha^{m,k}_i)}{\partial \alpha^{m,k}_i}\bigg]
\end{align*}
Please note that attributions are defined relative to an uninformative input called the baseline. We use a vector of all zeros as the baseline (denoted by $(\alpha^{m,k}_i)^{'}$).
Table~\ref{tab:clevr_ig} shows the results for this experiment.

The module structure parameters ($\bm{\alpha}$ parameters) of the \textit{Answer} modules have their attributions to the final probability around 1-2 orders of magnitudes higher than rest of the modules. 
The higher influence of Answer modules can be explained by the fact that they receive the memory features from the previous time-step and the classifier receives the memory features of the final time-step. The job of \textit{Attention} modules is to utilize intermediate attention maps to produce new feature maps which are used as input by the \textit{Answer} modules.

\begin{wrapfigure}[23]{r}{0.5\textwidth}
\begin{minipage}{0.5\textwidth}
\vspace{-5ex}
\resizebox{0.95\linewidth}{!}{
\centering\includegraphics[scale=0.3]{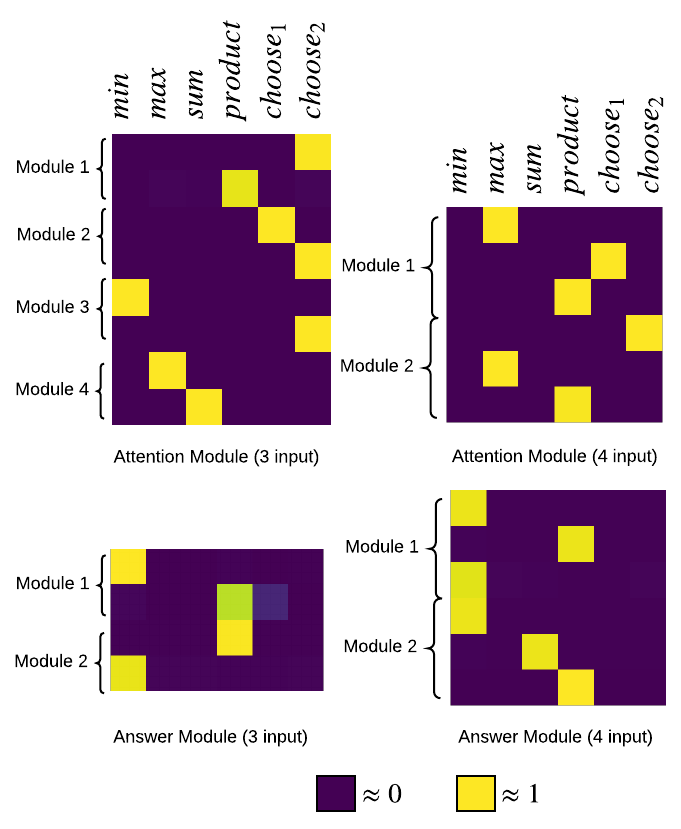}}
\caption{Visualization of module structure parameters (LNMN (11 modules)). For each module, each row denotes the $\bm{\alpha^{'}} = \bm{\sigma}(\bm{\alpha})$ parameters of the corresponding node.\label{fig:alpha_vis_70_11}}
\end{minipage}
\end{wrapfigure}
\subsection{Visualization of module network parameters}
In order to better interpret the individual contributions from each of the arithmetic operators to the modules, we plot them as color-maps for each type of module. The resulting visualizations are shown in Figure~\ref{fig:alpha_vis_70_11} for LNMN (11 modules). It is clear from the figure that the operation weights (or $\bm{\alpha^{'}}$ parameter) are approximately one-hot for each node. This is necessary in order to learn modules which act as composition of elementary operators on input feature maps rather than a mixture of operations at each node. The corresponding visualizations for LNMN (9 modules) and LNMN (14 modules) are given in Figure~\ref{fig:alpha_vis_70} and Figure~\ref{fig:alpha_vis_70_14} respectively (all of which are given in the Appendix~\ref{sec:vis_mod_appendix}).
 The analytical expressions of modules learned by LNMN (11 modules) are shown in Table~\ref{tab:module_eqn}. The diversity of modules as given in their equations indicates that distinct modules emerge from training.\\
\subsection{Measuring the role of individual arithmetic operators}
Each module (\textit{aka} cell) contains nodes which involves use of six elementary arithmetic operations (i.e. \textit{min}, \textit{max}, \textit{sum}, \textit{product}, \textit{choose\textunderscore1} and \textit{choose\textunderscore2}). We zero out the contribution to the node output for one of the arithmetic operations for all nodes in all modules and observe the degradation in the CLEVR validation accuracy\footnote{The CLEVR test set ground truth answers are not public, so we use the validation set instead. However, Table~\ref{tab:fullclevr} shows results for CLEVR test set (evaluated by the authors of CLEVR dataset).}. The results of this study are shown in Table~\ref{tab:clevr_rm_op}. 
The trend of overall accuracy shows that removing \textit{max} and \textit{product} operators results in maximum drop in overall accuracy ($\sim 50\%$). Other operators like \textit{min}, \textit{sum} and  \textit{choose\textunderscore1} result in minimal drop in overall accuracy. 
\begin{table*}[!ht]
\small
\centering
\resizebox{\linewidth}{!}{
\begin{tabular}{l|l}
\hline
Module type                           & Module implementation \\ \hline
\multirow{4}{*}{\specialcell{Attention\\(3 inputs)}} &       $O(img, a, c_{txt}) = conv_2(choose_2(conv_1(\mathcal{I}), a) \odot W_1 c_{txt}) = conv_2(a \odot W_1 c_{txt})$                \\
                                      &       $O(img, a, c_{txt})
= conv_2(choose_2(choose_1(conv_1(\mathcal{I}), a), W_1 c_{txt})) = conv_2(W_1 c_{txt})$                \\
                                      &       $O(img, a, c_{txt})
= conv_2(choose_2(min(conv_1(\mathcal{I}), a), W_1 c_{txt})) = conv_2(W_1 c_{txt})$                \\
                                      &       \specialcell{$O(img, a, c_{txt})
= conv_2(max(conv_1(\mathcal{I}), a) + W_1 c_{txt}))$}                \\ \hline
\multirow{2}{*}{\specialcell{Attention\\(4 inputs)}} &     \specialcell{$O(img, a_1, a_2, c_{txt}) = conv_2(choose_1(max(a_1, a_2), conv_1(\mathcal{I})) \odot W_1 c_{txt}))$\\$= conv_2(max(a_1, a_2) \odot W_1 c_{txt})$} \\
                                      &     \specialcell{$O(img, a_1,a_2,c_{txt})=conv_2(max(choose_2(a_1, a_2), conv_1(\mathcal{I})) \odot W_1 c_{txt}))$\\$= conv_2(max(a_2, conv_1(\mathcal{I})) \odot W_1 c_{txt}))$}\\ \hline
\multirow{2}{*}{\specialcell{Answer\\(3 inputs)}}                     &      \specialcell{$O(img, a, c_{txt})= W_2[\sum min(conv_1(\mathcal{I}), a) \odot W_1 c_{txt}, W_1 c_{txt}, f_{mem}]$}                 \\ &
\specialcell{$O(img, a, c_{txt})= W_2[\sum min((conv_1(\mathcal{I}) \odot a), W_1 c_{txt}),
W_1 c_{txt}, f_{mem}]$}\\ \hline
\multirow{2}{*}{\specialcell{Answer\\(4 inputs)}}                     &      \specialcell{$O(img, a_1, a_2, c_{txt})
= W_2[\sum min((min(a_1, a_2) \odot conv_1(\mathcal{I})), W_1 c_{txt}), W_1 c_{txt}, f_{mem}]$}                 \\ &
\specialcell{$O(img, a_1, a_2, c_{txt})
= W_2[\sum((min(a_1, a_2) + conv_1(\mathcal{I})) \odot W_1 c_{txt}), W_1 c_{txt}, f_{mem}]$}                 \\ \hline
\end{tabular}}
\caption{Analytical expression of modules learned by LNMN (11 modules). In the above equations, $\sum$ denotes sum over spatial dimensions of the feature tensor.}
\label{tab:module_eqn}
\end{table*}




\begin{table*}[!htbp]
\small
\centering
\begin{tabular}{c|c|cccccc}
\toprule
\specialcell{Operator\\Name}   & Overall & Count & Exist & \specialcell{Compare\\number} & \specialcell{Query\\attribute} & \specialcell{Compare\\Attribute} \\ \midrule
min       & 86.64 &	77.98 &	86.79 &	87.89 &	88.77 &	93.10              \\
max       & 45.54 &	35.92 &	55.25 &	63.66 &	40.52 &	51.83              \\
sum       & 82.67 &	69.89 &	80.25 &	85.22 &	87.69 &	90.05              \\
product       & 34.65 &	14.55 &	51.49 &	48.79 &	30.31 &	49.92              \\
choose\textunderscore 1 & 89.74 &	84.24 &	93.81 &	89.02 &	89.59 &	94.67              \\
choose\textunderscore 2 & 79.45 &	64.77 &	76.07 &	82.96 &	86.78 &	84.94             \\ \hline
Original Model & 89.88 & 84.28 & 93.74 & 89.63 & 89.64 & 94.84  \\ \bottomrule
\end{tabular}
\caption{Analysis of performance drop with removing operators from a trained model (LNMN 9 modules) on CLEVR validation set.}
\label{tab:clevr_rm_op}
\end{table*}

\section{Related Work}\label{related_work}

\textbf{Neural Architecture Search}: Neural Architecture Search (NAS) is a  technique to automatically learn the structure and connectivity of neural networks rather than training human-designed architectures. In \cite{zoph2016neural}, a recurrent neural network (RNN) based controller is used to predict the hyper-parameters of a CNN such as number of filters, stride, kernel size etc. They used REINFORCE \cite{williams1992simple} to train the controller with validation set accuracy as the reward signal.
As an alternative to reinforcement learning, evolutionary algorithms \cite{stanley2017neuroevolution} have been used to perform architecture search in \cite{real2017large, miikkulainen2019evolving, liu2017hierarchical, real2018regularized}. Recently, \cite{liu2018darts} proposed a differentiable approach to perform architecture search and reported success in discovering high-performance architectures for both image classification and language modeling. \cite{kirsch2018modular} proposes an EM style algorithm to learn black-box modules and their layout for image recognition and language modeling tasks.

\textbf{Visual Reasoning Models}: Among the end-to-end models for the task of visual reasoning, FiLM \cite{DBLP:journals/corr/abs-1709-07871} uses Conditional Batch Normalization (CBN) \cite{de2017modulating,dumoulin2017learned} to modulate the channels of input convolutional features in a residual block. \cite{hudson2018compositional} obtains the features by iteratively applying a Memory-Attention-Control (MAC) cell that learns to retrieve information from the image and aggregate the results into a recurrent memory. \cite{santoro2017simple} constructs the feature representation by taking into account the relational interactions between objects of the image. With regards to the modular approaches, \cite{andreas2016neural} proposes to compose neural network modules (with shared parameters) for each input question based on layout predicted by syntactic parse of the question. \cite{andreas2016learning} extends this approach to question-answering in a database domain. In \cite{hu2017learning}, the layout prediction is relaxed by learning a layout policy with a sequence-to-sequence RNN. This layout policy is jointly trained along with the parameters of modules. In \cite{johnson2017inferring}, the modules are residual blocks (convolutional), they learn the program generator separately and then fine-tune it along with the modules. TbD-net \cite{Mascharka2018TransparencyBD} builds upon the End-to-End Module Networks \cite{hu2017learning} but makes an important change in that the proposed modules explicitly utilize attention maps passed as inputs instead of learning whether or not to use them. This results in more interpretability of the modules since they perform specific functions.

\textbf{Visual Question Answering}: Visual question answering requires a learning model to answer sophisticated queries about visual inputs.
Significant progress has been made in this direction to design neural networks that can answer queries about images. This can be attributed to the availability of relevant datasets which capture real-life images like DAQUAR ~\cite{malinowski2014multi}, COCO-QA \cite{ren2015exploring} and most recently VQA (v1 \cite{VQA} and v2 \cite{balanced_vqa_v2}). 
The most common approaches \cite{ren2015image, noh2016image} to this problem include construction of a joint embedding of question and image and treating it as a classification problem over the most frequent set of answers. Recent works \cite{jabri2016revisiting,johnson2017clevr} have shown that the neural networks tend to exploit biases in the datasets without learning how to reason.


\section{Conclusion}\label{conclusion}
We have presented a differentiable approach to learn the modules needed in a visual reasoning task automatically. With this approach, we obtain results comparable to an analogous model in which modules are hand-specified for a particular visual reasoning task. In addition, we present an extensive analysis of the degree to which each module influences the prediction function of the model, the effect of each arithmetic operation on overall accuracy and the analytical expressions of the learned modules. In the future, we would like to benchmark this generic learnable neural module network with various other visual reasoning and visual question answering tasks. 

\FloatBarrier




\bibliographystyle{plain}
\bibliography{main}

\newpage

\appendix
\section{Appendix}
\label{sec:appendix}


\begin{algorithm}
\caption{Operation of a module}\label{alg:run_module}
\begin{algorithmic}
\Procedure{run-module}{$m, A, p, \bm{c_t}, \mathcal{I}$}
\State $a_1 \gets \sum_{i=1}^L A_i \cdot p_i$\Comment{Read from stack}
\State $p \gets$ \textrm{1D-conv}$(p, [0, 0, 1])$\Comment{decrement the stack pointer}\\
\If{no. of inputs==4}
{
\State $a_2 \gets \sum_{i=1}^L A_i \cdot p_i$\Comment{Read from stack}
\State $p \gets \textrm{1D-conv}(p, [0, 0, 1])$\Comment{decrement the stack pointer}
\State $o_m \gets m(\mathcal{I}, \bm{c_t}, a_1, a_2)$
}
\Else{\State $o_m \gets m(\mathcal{I}, \bm{c_t}, a_1)$}
\State $p \gets$ 1D-conv$(p, [1, 0, 0])$\Comment{increment the stack pointer}\\
\For{$i = 1,...,L$}{
\State $A \gets$ $A\cdot(1-p_i) + o_m \cdot p_i$\Comment{Write to stack}}
\Return $A$, $p$
\EndProcedure
\end{algorithmic}
\label{algo:module}
\end{algorithm}

\begin{figure}[!ht]
\centering\includegraphics[scale=0.45]{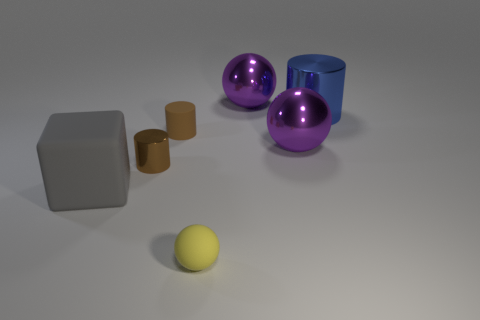}
\caption{\textit{\textbf{Q1}: What number of cylinders are gray objects or tiny brown matte objects? \textbf{A}: 1\\ \textbf{Q2}: Is the number of brown cylinders in front of the brown matte cylinder less than the number of brown rubber cylinders? \textbf{A}: no}\label{fig:clevr_img}}
\end{figure}

\begin{figure}[!ht]
\centering\includegraphics[scale=0.6]{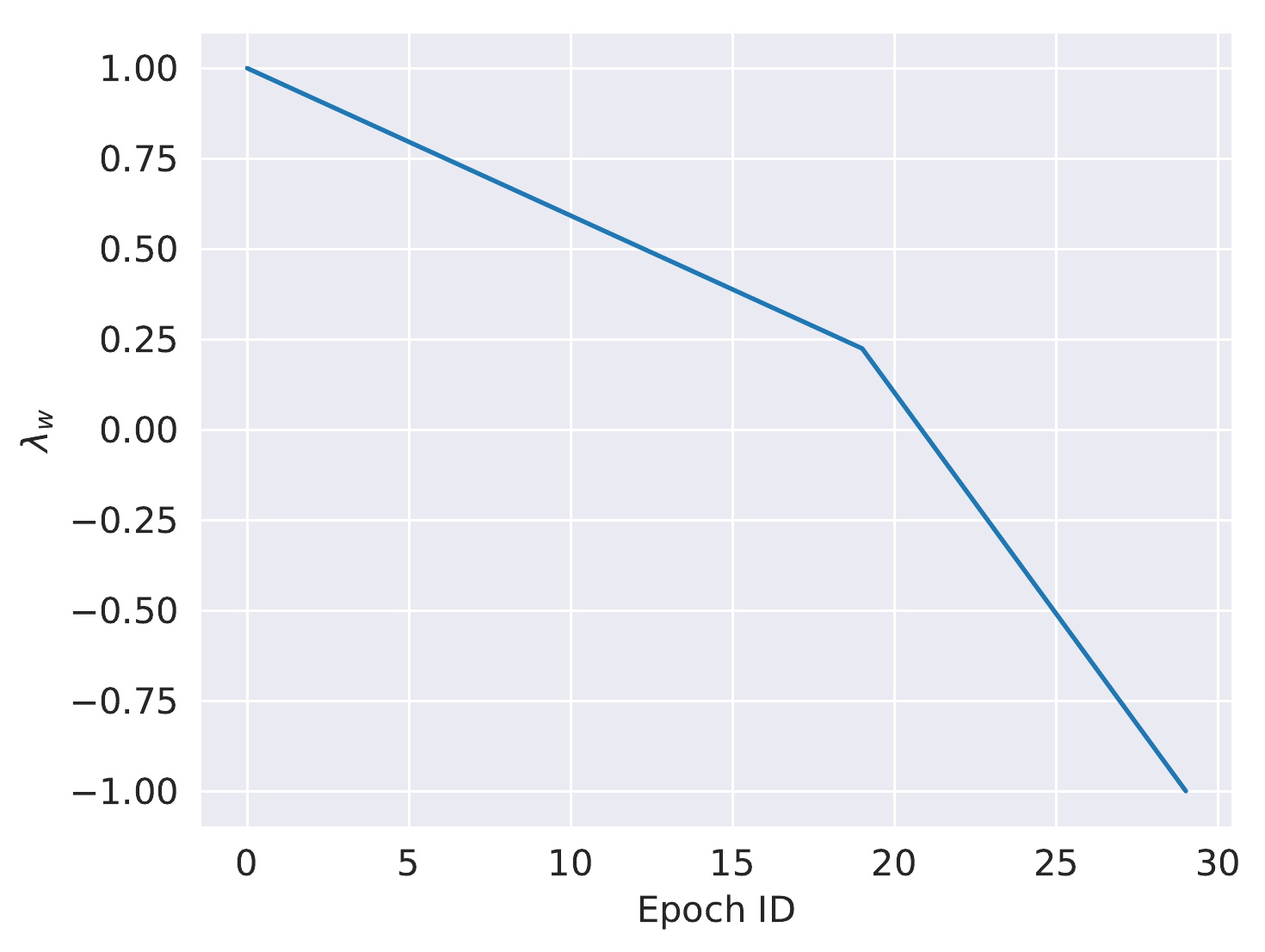}
\caption{Plot of variation of $\lambda_w$ with epochs.}\label{fig:reg_coeff_plot}
\end{figure}

\FloatBarrier
\subsection{Module schematic diagrams}\label{sec:module_diag_appendix}

\begin{figure*}[!ht]
\centering\includegraphics[scale=0.3]{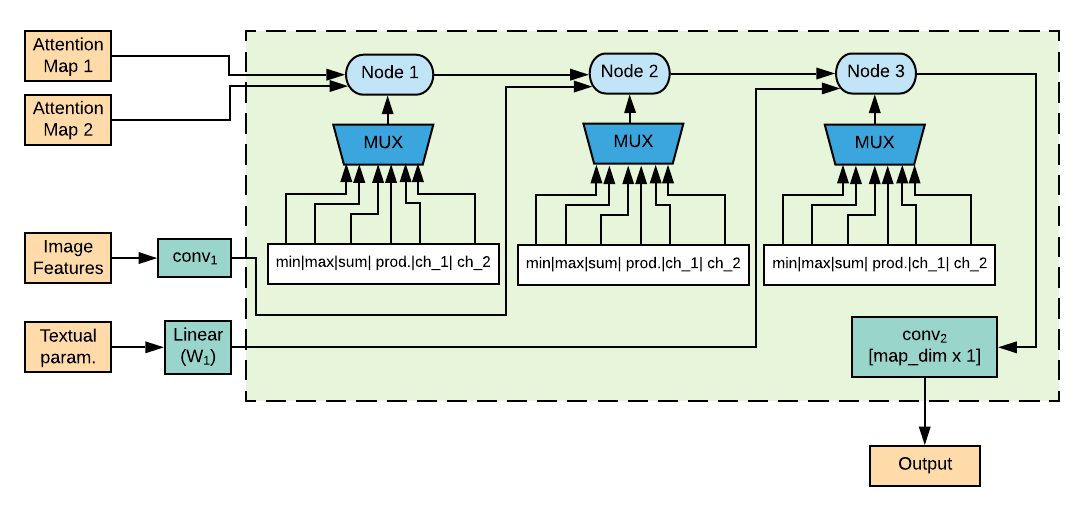}
\caption{Attention Module schematic diagram (4 inputs).\label{fig:cell_structure_4_input}}
\vspace{-2ex}
\end{figure*}
\begin{figure*}[!ht]
\centering\includegraphics[scale=0.3]{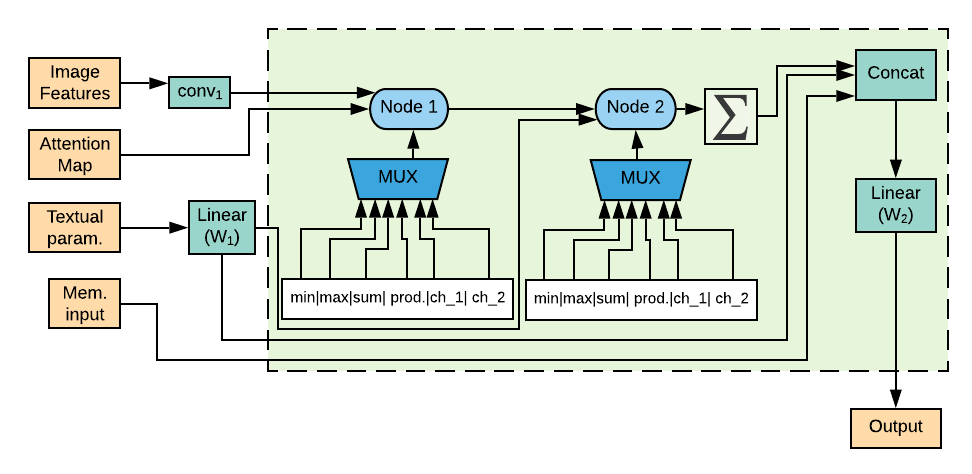}
\caption{Answer Module schematic diagram (3 inputs)\label{fig:cell_structure_3_input_answer}}
\end{figure*}

\begin{figure*}[!ht]
\centering\includegraphics[scale=0.3]{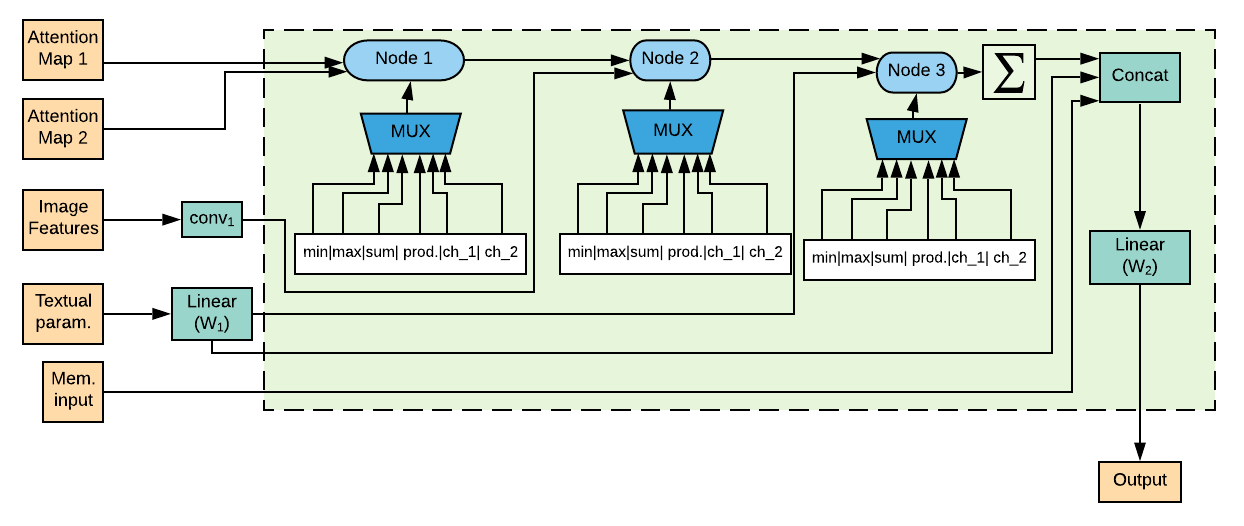}
\caption{Answer Module schematic diagram (4 inputs)\label{fig:cell_structure_4_input_answer}}
\end{figure*}

\FloatBarrier
\subsection{Hand-crafted modules of Stack-NMN}\label{sec:hand_modules}

\begin{table}[!htbp]
\centering
\begin{tabular}{|l|c|c|c|l|}
\hline
module & input & output & implementation details \\
name & attention & type & ($x$: image feature map, $c$: textual parameter) \\
\hline
\texttt{Find} & (none) & attention & $a_{out}=\mathrm{conv_2}\left(\mathrm{conv_1}(x) \odot W c\right)$ \\
\texttt{Transform} & $a$ & attention & $a_{out}=\mathrm{conv_2}\left(\mathrm{conv_1}(x) \odot W_1\sum(a \odot x) \odot W_2 c\right)$ \\
\texttt{And} & $a_1, a_2$ & attention & $a_{out} = \mathrm{minimum}(a_1, a_2)$ \\
\texttt{Or} & $a_1, a_2$ & attention & $a_{out} = \mathrm{maximum}(a_1, a_2)$ \\
\texttt{Filter} & $a$ & attention & $a_{out} = \mathtt{And}(a, \mathtt{Find}())$, i.e. reusing \texttt{Find} and \texttt{And} \\
\texttt{Scene} & (none) & attention & $a_{out}=\mathrm{conv_1}(x)$ \\
\texttt{Answer} & $a$ & answer & $y=W_1^T \left(W_2\sum(a \odot x) \odot W_3 c\right)$ \\
\texttt{Compare} & $a_1, a_2$ & answer & $y=W_1^T \left(W_2\sum(a_1 \odot x) \odot W_3\sum(a_2 \odot x) \odot W_4 c\right)$ \\
\texttt{NoOp} & (none) & (none) & (does nothing) \\
\hline
\end{tabular}
~\\~\\
\caption{Neural modules used in \cite{hu2018explainable}. The modules take image attention maps as inputs, and output either a new image attention $a_{out}$ or a score vector $y$ over all possible answers ($\odot$ is elementwise multiplication; $\sum$ is sum over spatial dimensions).}
\label{tab:modules}
\end{table}

\FloatBarrier

\newpage
\subsection{Visualization of module structure parameters}\label{sec:vis_mod_appendix}
\begin{figure}[!ht]
\centering\includegraphics[scale=0.32]{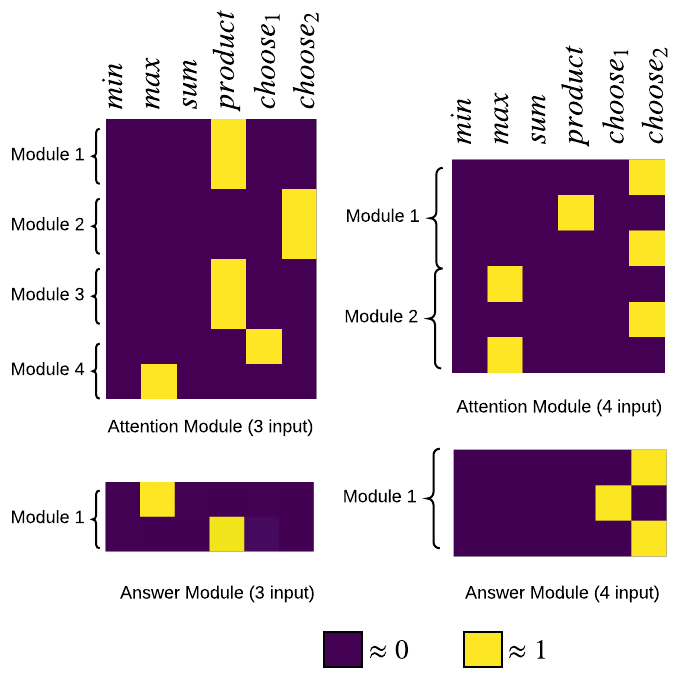}
\caption{Visualization of module structure parameters (LNMN (9 modules)). For each module, each row denotes the $\bm{\alpha^{'}} = \bm{\sigma}(\bm{\alpha})$ parameters of the corresponding node.\label{fig:alpha_vis_70}}
\end{figure}



\begin{figure}[!ht]
\centering\includegraphics[scale=0.32]{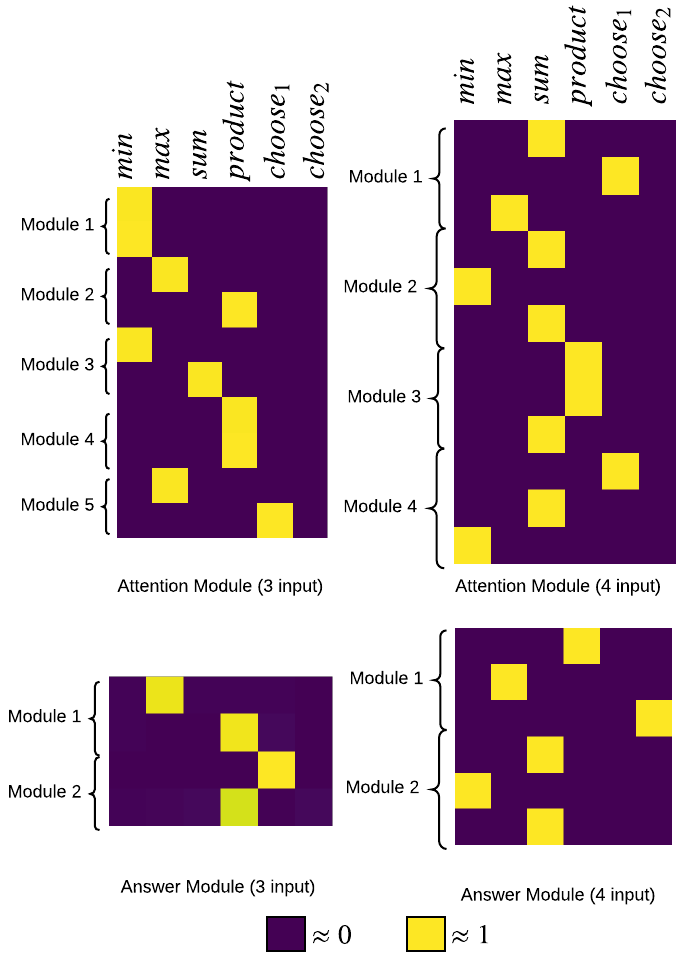}
\caption{Visualization of module structure parameters (LNMN (14 modules)). For each module, each row denotes the $\bm{\alpha^{'}} = \bm{\sigma}(\bm{\alpha})$ parameters of the corresponding node.\label{fig:alpha_vis_70_14}}
\end{figure}

\FloatBarrier
\subsection{Results on Natural Image VQA datasets}\label{sec:acc_vqa}

\begin{table}[!ht]
\centering
\begin{tabular}{ccccc}
\toprule
Model     & Overall & Yes/No & Number & Other \\ \midrule
Stack-NMN & 59.84   & 80.75  & 37.49  & 46.83 \\
LNMN (9 modules)  & 57.67   & 80.41  & 36.65  & 42.82 \\
\bottomrule
\end{tabular}
\caption{Test Accuracy on VQA v1 \cite{VQA}}
\label{tab:results_vqa_1}
\end{table}

\begin{table}[!ht]
\centering
\begin{tabular}{ccccc}
\toprule
Model            & Overall & Yes/No & Number & Other \\ \midrule
Stack-NMN        & 58.23   & 77.06  & 37.48  & 46.59 \\
LNMN (9 modules) & 54.85   & 73.78  & 35.05  & 42.92 \\
\bottomrule
\end{tabular}
\caption{Test Accuracy on VQA v2 \cite{balanced_vqa_v2}}
\label{tab:results_vqa_2}
\end{table}





\end{document}